\documentclass[lettersize,journal]{IEEEtran}
\usepackage{amsmath,amsfonts}
\usepackage{algorithmic}
\usepackage{algorithm}
\usepackage{array}
\usepackage[caption=false,font=normalsize,labelfont=sf,textfont=sf]{subfig}
\usepackage{textcomp}
\usepackage{stfloats}
\usepackage{xurl}

\usepackage{verbatim}
\usepackage{graphicx}
\usepackage{cite}
\newcolumntype{P}[1]{>{\centering\arraybackslash}p{#1}}

\begin{document}

\title{Graph-Powered Defense: Controller Area Network Intrusion Detection for Unmanned Aerial Vehicles}

\author{ Reek Majumder, Ph.D. ~\IEEEmembership{Member,~IEEE,}, Gurcan Comert, Ph.D.~\IEEEmembership{Member,~IEEE,}, David Werth, Ph.D.~\IEEEmembership{Member,~IEEE,}, Adrian Gale, Ph.D.~\IEEEmembership{Member,~IEEE,}, Mashrur Chowdhury, Ph.D., P.E.~\IEEEmembership{Senior Member,~IEEE,}, M Sabbir Salek, Ph.D. ~\IEEEmembership{Member,~IEEE,}
% IEEE Publication Technology,~\IEEEmembership{Staff,~IEEE,}
        % <-this % stops a space
}

\markboth{IEEE INTERNET OF THINGS (IoT) JOURNAL}%
{Majumder \MakeLowercase{\textit{et al.}}: Graph-Powered Defense: Controller Area Network Intrusion Detection for Unmanned Aerial Vehicles}



\maketitle

\begin{abstract}
The network of services, including delivery, farming, and environmental monitoring, has experienced exponential expansion in the past decade with Unmanned Aerial Vehicles (UAVs). Yet, UAVs are not robust enough against cyberattacks, especially on the Controller Area Network (CAN) bus. The CAN bus is a general-purpose vehicle-bus standard to enable microcontrollers and in-vehicle computers to interact, primarily connecting different Electronic Control Units (ECUs). In this study, we focus on solving some of the most critical security weaknesses in UAVs by developing a novel graph-based intrusion detection system (IDS) leveraging the Uncomplicated Application-level Vehicular Communication and Networking (UAVCAN) protocol. First, we decode CAN messages based on the UAVCAN protocol specification; second, we present a comprehensive method of transforming tabular UAVCAN messages into graph structures. Lastly, we apply various graph-based machine learning models for detecting cyber-attacks on the CAN bus, including graph convolutional neural networks (GCNNs), graph attention networks (GATs), Graph Sample and Aggregate Networks (GraphSAGE), and graph structure-based transformers. Among these, GCNNs operate in a transductive setting, requiring access to the full graph structure during both training and inference. In contrast, GraphSAGE, graph transformers and GATs follow an inductive learning method, where the models learns to generate node embeddings by sampling and aggregating neighborhood information. This enables them to generalize effectively to previously unseen nodes or entirely new graph instances, making them suitable for real-world CAN bus environments. Our findings show that inductive models can achieve competitive and even better accuracy than transductive models in detecting various types of intrusions, with minimum information on protocol specification, thus providing a generic robust solution for CAN bus security for the UAVs. We also compared our results with baseline single-layer Long Short-Term Memory (LSTM) and found that all our graph-based models perform better without using any decoded features based on the UAVCAN protocol, highlighting higher detection performance with protocol-independent capability. 
\end{abstract}


\section{Introduction}
\IEEEPARstart{U}{nmanned} Aerial Vehicles (UAVs), and their commercial variants, drones, have seen phenomenal growth in the past decade, with their application becoming ubiquitous, seemingly in every domain, from product delivery to agriculture, surveillance, and environmental monitoring. Deployment of UAVs with diverse sensors and communication technology provides significant business benefits, such as cost reduction, efficiency improvement, and capability for many different commercial sectors. By 2026, in the United States alone, drone applications will have an annual impact on the country’s GDP in the range of USD 31 billion to USD 46 billion, driven by innovations in UAV technology and developments in application areas \cite{ref1}.

Electronic Control Units (ECUs) are the main vital component in the modern UAV system. The main responsibilities of ECUs include controlling and managing the electrical subsystems of a UAV. For instance, ECUs are considered the brains of UAVs, processing data from sensors and executing commands to ensure smooth and safe operations. Most of the ECUs used in such systems are connected through the common Controller Area Network (CAN) bus, which is one of the well-known vehicle bus standards that is widely used in automotive and aerospace applications \cite{ref2,ref3}. The CAN bus facilitates communication between microcontrollers and devices without host computers. It implements strategies to control networking issues like error detection, data integrity, and data consistency \cite{ref4}. So, the CAN bus provides simple, efficient, and cost-effective solutions for networking in UAVs by enabling real-time communications between UAV modules, controlled feedback, and monitoring critical systems \cite{ref5}.

The security of CAN bus networks is critical, particularly as UAVs are increasingly integrated into a complex and potentially hostile environment. However, the CAN bus provides low latency and reliable network communication features. Still, it lacks integrated security features like authentication, authorization, data encryption, and network segmentation, which could make it vulnerable in potentially hostile UAV environments \cite{ref6}. With UAVs now integrated with more complex environments, this security gap could allow them to be compromised, especially if the ECUs have to communicate outside the network to gain more situational awareness \cite{ref7,ref8,ref9}. Previous research has highlighted the need for embedded security frameworks inside in-vehicle networks such as CAN and highlighted how CAVs, including UAVs, increase the attack surface through their cyber-physical integration \cite{ref10}.

Several vulnerabilities have been documented, such as command injection attacks via DJI universal markup language (DUML) on DJI drones, which include about 94\% of the world’s consumer and commercial drones. During this attack, the attacker exploits DUML commands to disrupt the communication between the drone and the Remote Controller (RC) communicating via the OcuSync protocol \cite{ref11}. Furthermore, commercial delivery drones have recently received a Beyond Visual Line of Sight (BVLOS) clearance from the Federal Aviation Administration (FAA) for their MK27 aerial drone \cite{ref12}. This could make drones more vulnerable to cyberattacks, where attackers could potentially gain control of a drone’s navigation either by signal jamming or GPS spoofing to divert it from its intended path or steal it \cite{ref13}. For example, in a high-profile incident, Iran intercepted and gained control of a US RQ-170 Sentinel stealth drone with BVLOS capability. By jamming its GPS and exploiting navigation vulnerability, Iranian forces mislead the UAV, causing it to land within Iranian territory \cite{ref14}. These examples exemplify the need for embedded intrusion detection systems (IDS) to detect and prevent unwanted access in real-time, reducing the risk of security breaches. 

UAV networks are becoming increasingly dependent on seamless connectivity, making them susceptible to attacks such as GPS-spoofing, jamming, malware injection, and denial of service (DoS) during dynamic drone missions or multi-drone operations. The cost of neglecting cybersecurity measures in UAVs might have serious economic and social risks. For example, a compromised drone in a commercial sector, specifically in a supply chain application, could disrupt operations and lead to financial loss. Some vulnerabilities include attacks on crucial ECUs like Flight Controller Units (FCUs) via CAN bus. At the same time, some are linked to ground-based control systems and external communication, which can increase attack surfaces for UAVs, compromising the safety and functionality of the UAVs. 

\begin{figure*}[!t]
\centering
\includegraphics[width=0.8\textwidth]{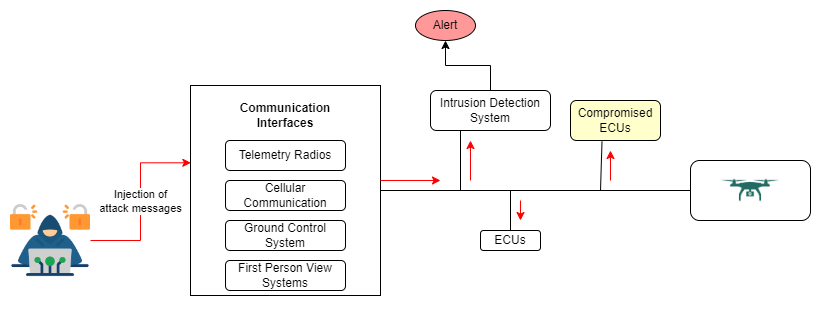}
\caption{Intrusion detection system for in-vehicle network security}
\label{fig:fig_1}
\end{figure*}

Enhancing threat detection through IDS in UAVs can generate a strong return on investment (ROI) by reducing the risk of mission failure, data breaches, and operational downtime, representing significant cost savings \cite{ref15}. In sectors like drone delivery, where large e-commerce companies are focusing on using drones to improve delivery time, an attack on these operational drones will lead to disruption of fleet operations, delayed deliveries, damage to their reputation, and lost revenue.  In defense applications, compromised UAVs will lead to classified data breaches, potentially costing the government millions and compromising national security \cite{ref16}. 

The following steps are crucial as they need collaboration from the manufacturers of UAVs, the cybersecurity specialists, and the government regulatory bodies. Firstly, the manufacturers must focus on designing UAVs with IDS capabilities so that high-performance intrusions, such as message injection and jamming, can be detected in real-time without additional hardware \cite{ref17}. Secondly, cybersecurity experts need to work closely with the manufacturers to update deployed IDS systems and make them adaptable to new threats. Thirdly, regulatory agencies, such as the FAA and the European Union Aviation Safety Agency (EASA), should define cybersecurity standards for consumer and commercial UAVs to ensure deployed drones have sufficient security resilience. These steps will help minimize the financial losses that might occur due to cyberattacks on drones and improve their reliability for multiple domains \cite{ref15}.

For a cyber-secure future for drones, extensive research is being conducted to create IDSs based on numerous features that have been proposed to safeguard UAV’s CAN bus networks \cite{ref18}. \cite{ref18} also suggests that the cost associated with integrating robust IDS is much less than the financial loss of a cyberattack that may lead to serious data breaches, security, and privacy issues. The integration of an IDS into UAV communication networks is depicted in Figure~\ref{fig:fig_1}. It illustrates how IDS can work against injected attack messages aimed at compromising communication interfaces and ECUs. This configuration exhibits how the IDS safeguards the network and maintains operational integrity by guarding against possible ECU compromises.

There are two major issues with currently available IDSs; first, some solutions require understanding the details provided by the CAN protocol or altering it by adding some authentication field. This becomes a challenge for many modern vehicles, such as UAVs, that use closed-source protocols, making these solutions impractical to apply in the real world. Secondly, some IDSs use advanced machine learning models that require large amounts of carefully labeled CAN messages, including attack and non-attack scenarios. Compiling data for various attacks can be challenging in generating an effective classifier \cite{ref5,ref19}. 
% Our method aims to overcome these issues by proposing a CAN protocol-independent but density-based parameter-monitoring system.

We tackle both these challenges by introducing a CAN protocol-independent intrusion detection method. By protocol-independent, we mean that our method does not need access to the inner structure of the CAN data payload as specified by a protocol (for example, UAVCAN). Although our dataset was collected from a CAN-based system, our method only requires structural characteristics, e.g., the ordering of CAN IDs to build a temporal graph. From these graphs, node-level features like PageRank are extracted to detect anomalies, and thus, no payload-specific decoding is needed.

In addition, our method uses density-based parameter monitoring to characterize normal runtime distributions of graph-based features to detect anomalies as potential intrusions. This makes it lightweight and more generally applicable to systems that use different, even proprietary, CAN extensions or lack full protocol-level access. Moreover, implementing advanced IDS may include specific hardware such as a hardware security module (HSM), to support a secure boot process, or a processor with cryptographic acceleration capability like AURIX TC399, which can support data encryption. 

Adding secure boot or specialized hardware like AURIX TC399 can significantly increase the cost of UAV systems \cite{ref4}, though such investment may be justified for safety-critical missions or when sensitive data is at risk \cite{ref20}. As a cost-effective alternative, we explore graph-based IDS models alongside a baseline Long Short-Term Memory (LSTM)-based model for the UAVCAN protocol. Unlike deep learning-based IDS that depend on GPUs like Nvidia Jetson Nano or Xavier\cite{ref21}, graph-based models can run efficiently on lightweight, energy-efficient microcontrollers such as ARM Cortex-M \cite{ref22}. Some, like GraphSAGE, have even been deployed on FPGA-CPU setups with low latency and no need for GPUs \cite{ref23}.

We hypothesize that utilizing graph-structured data can enhance the IDSs by leveraging the structural properties of a graph, thus making communication through the CAN bus more resilient to injection attacks. This approach enables effective detection without relying on costly or computationally intensive hardware.

\section{Contribution}

This study introduces a lightweight, protocol-independent, and hardware-efficient intrusion detection framework for UAVs by leveraging structural patterns in CAN bus communication graphs to detect injection attacks. Instead of analyzing CAN message contents, our approach focuses on the structure and flow of messages, eliminating the need to decode payloads or rely on specialized hardware. This design enhances adaptability and makes the framework well-suited for resource-constrained environments. The primary contributions of this study are as follows:

\begin{itemize}
    % \item A protocol-independent IDS framework for CAN-based UAV systems that detects anomalies without decoding the message payload, making it suitable for systems with limited or proprietary protocol access.
    % \item We present a protocol-independent intrusion detection system (IDS) framework for CAN-based UAV systems that identifies anomalies without decoding message payloads of CAN, making it practical for applications with limited or proprietary protocol access.
    \item \textbf{Protocol-Independent IDS Framework:} We propose a protocol-agnostic intrusion detection system (IDS) for CAN-based UAV platforms that identifies anomalous behavior without decoding the CAN message payload. This makes the framework applicable even in scenarios with limited access to protocol specifications or proprietary message formats.
    
    % \item We propose a novel, generalizable graph construction strategy that captures temporal and structural patterns from raw CAN bus traffic by incorporating message sequencing, transfer ID-based self-loops, and timestamp-based edge weighting, enabling the generation of dynamic communication graphs, which are used to model communication behavior for intrusion detection.
    % \item A method to construct dynamic, weighted graphs from raw CAN bus traffic, capturing communication behavior using only message IDs and timing information.
    \item \textbf{Novel Graph Construction Strategy:} We introduce a generalizable graph construction method that captures temporal and structural patterns from raw CAN bus data. By incorporating message sequencing, transfer ID-based self-loops, and timestamp-based edge weighting, we generate dynamic communication graphs that effectively model the communication behavior of CAN bus data for detecting intrusions.
    
    % \item The development and evaluation of multiple graph-based machine learning models, including graph convolutional neural networks (GCNNs), graph attention networks (GATs), Graph Sample and Aggregate Networks (GraphSAGE), and graph structure-based transformers to detect injection attacks in CAN network.
    % \item We compare our graph-based models with a baseline LSTM-based IDS and show that they achieve better detection performance, even without accessing the message payload—unlike the LSTM model, which relied on payload decoding to function effectively.
    \item \textbf{Performance Evaluation Against Payload-Dependent Models:} We evaluate our graph-based IDS against a baseline LSTM-based model that relies on message payload decoding. Our approach demonstrates superior detection performance despite operating without access to payload data, highlighting its robustness and practicality of our method in real-world UAV deployments.
\end{itemize}

\section{Literature Review}

UAVs are evolving from simple remote-controlled aerial devices to fully autonomous systems that rely on complex networks of sensors and controllers for their operations. Integrating asynchronous half-duplex communication protocols, such as CAN buses in UAVs, makes them more vulnerable to cyberattacks than those targeting automotive systems. Thus, it emphasizes the need for more robust and reliable IDS systems for UAV networks. Though both UAVs and autonomous systems use the CAN bus protocol for reliable and fast real-time communication, the application scenarios provide some key differences. Automobile CAN buses often interoperate with additional networks such as LIN \cite{ref24} or FlexRay \cite{ref25} to process hierarchical data, minimize cost and improve performance. On the other hand, UAVs focus on a centralized and compact CAN bus structure to address stringent dimensions, weight and energy requirements. In addition, security concerns differ significantly because automotive CAN buses are typically confined to a vehicle and have security focused on the vehicle, while UAVs use wireless communications channels for remote operation, leaving their CAN bus networks vulnerable to a broader spectrum of remote attacks \cite{ref26}. Thus, it emphasizes the need for more robust and reliable IDS systems for UAV networks.

\begin{figure*}[!t]
\centering
\includegraphics[width=0.8\textwidth]{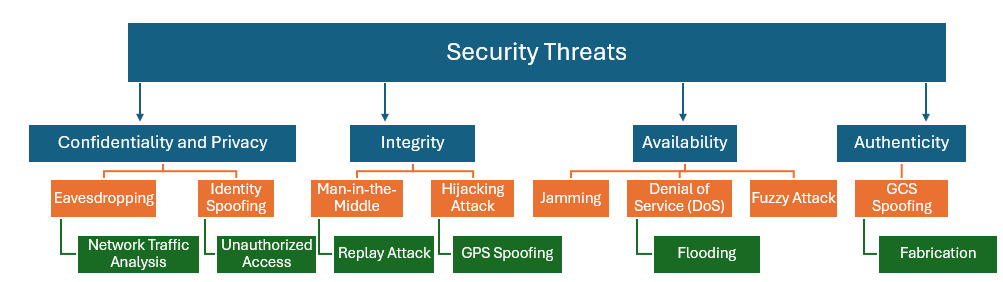}
\caption{Security threats to CAN bus network}
\label{fig:fig_2}
\end{figure*}

A CAN provides a dedicated, dependable, and effective communication channel for all in-vehicle-connected ECUs, sensors, and systems. It is widely accepted as a standard for in-vehicle communication systems. Recent studies have shown that CAN is vulnerable to various cyberattacks such as denial of service (DoS), spoofing, fuzzy, flooding and replay attacks \cite{ref27,ref28} due to a lack of security measures like authentication, authorization, and encryption schemes within the protocol \cite{ref29,ref30,ref31}. These attacks compromise the network’s confidentiality, integrity, and availability by introducing fake messages to the CAN bus. Figure~\ref{fig:fig_2} depicts several possible security threats to UAV ECUs communicating through the CAN bus interface.

% \begin{figure}[!t]
% \centering
% \includegraphics[width=3.5in]{pic2_view3.png}
% \caption{Security threats to CAN bus network}
% \label{fig_2}
% \end{figure}
% % \vspace{-15pt}

Techniques like network segmentation, encryption, and authentication have been proposed to enhance the security of the CAN bus, thus mitigating attacks through vehicular networks. These methods will stop attacks by automatically initiating predefined policies against unknown threats \cite{ref32,ref33}. Network segmentation improves CAN bus security by creating subnetworks, which control access to specific segments of the network \cite{ref34}. Another method includes encrypting data from the CAN frame using a dynamically changing symmetric key generator \cite{ref35}. Some research efforts have combined encryption and authentication of CAN messages using the stream cipher RC4 \cite{ref36}. However, integrating authentication, authorization, and encryption solutions poses a significant challenge, especially when modern vehicles demand real-time communication. To address this drawback, other IDS were investigated \cite{ref37,ref38}. The IDS for in-vehicle networks can be divided into four types: fingerprint-based \cite{ref39,ref40}, parameter-monitoring-based \cite{ref41,ref42,ref43}, information-theoretic-based \cite{ref44}, and machine-learning-based systems \cite{ref45,ref46}. Our proposed method falls into the parameter-monitoring and machine learning-based systems class, where we track the density \cite{ref43} of a particular CAN ID and use it as a feature for graph-based machine learning (GB-ML) models.

In recent years, the emergence of new types of malware has accelerated alarmingly. This trend highlights the necessity for an algorithm to handle future unknown attacks, as signature-based methods are only effective against known attacks. Researchers have presented various machine-learning and deep-learning-based IDS. In \cite{ref47}, support vector machine-based classification of regular and anomalous CAN frames was proposed. Similarly, \cite{ref48} uses deep neural network-based packet classification. Another method, based on time-series analysis using a Long Short-Term Memory (LSTM) based IDS \cite{ref45}, showed a performance of over 87\% for potent replay attacks. LSTM approaches were studied in depth in Machine Learning-based Intrusion Detection Systems (MLIDS) in \cite{ref49}, and it was found that these models can slow down execution in a low-resource environment, specifically when considering UAVs with other sensors. A study evaluated the quantum-restricted Boltzmann machine \cite{ref46} and hybrid quantum-classical neural network \cite{ref50} for CAN intrusion detection, leveraging advancements in quantum technology to improve detection capabilities. However, these methods will still take some time to be implemented in real-time due to resource constraints in quantum machines \cite{ref51,ref52}.

Recently, research on GB-ML models has drawn much attention because of the use of graphs in large language models \cite{ref53}. The latter can be designed to store knowledge about related data items and their context, and GB-ML can be used to harness that information. Some graph-based anomaly detection strategies incorporate evaluation metrics as background knowledge. For example, in \cite{ref54} background knowledge is added as a rule coverage reporting the percentage of the final graph covered by the instances of the substructure. They hypothesize that anomalous structures can be detected by giving negative weights to rules. In \cite{ref55}, the authors present graph-based outlier detection using real-time home IoT traffic as graph streams to detect DoS attacks while processing graph data in real-time.

Besides anomaly detection, GB-ML techniques have also been applied to enhance IDSs in various network infrastructures, such as the CAN bus in automobiles. In \cite{ref6}, a graph-based approach was combined with a variational autoencoder (VAE) to train the classifier on positive samples only. Similarly, a graph convolution neural network (GCNN)  to analyze CAN data was utilized in \cite{ref56}. This approach minimized the need for extensive feature engineering and excelled in detecting mixed attacks, such as combinations of DoS, fuzzy, spoofing, and replay attacks.

Despite significant advances in GB-ML, there is a significant lack of comprehensive research applying it to real-world, large-scale systems. The potential for generalization of GB-ML approaches across attacks and environments remain largely underexplored. This creates a promising avenue for future research that could lead to robust, scalable, and adaptive IDSs capable of securing critical systems.

\section{Background}
This section covers the introduction of the CAN bus and the security flaws discovered in the CAN bus. We also discussed the UAVCAN protocol built by the UAV community to improve drone network security. Furthermore, we discuss various attacks on the UAVCAN protocol via the CAN bus.

\begin{figure*}[!t]
\centering
\subfloat[Standard CAN 2.0A frame]{\includegraphics[width=5.5in]{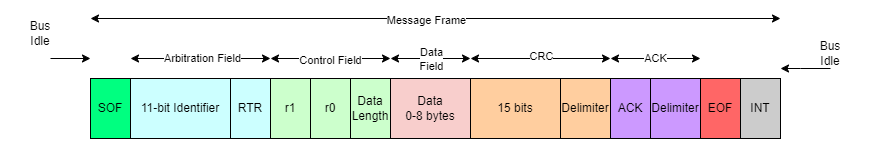}%
\label{fig_standard_can}}
\hfil
\subfloat[Extended CAN 2.0B frame]{\includegraphics[width=5.5in]{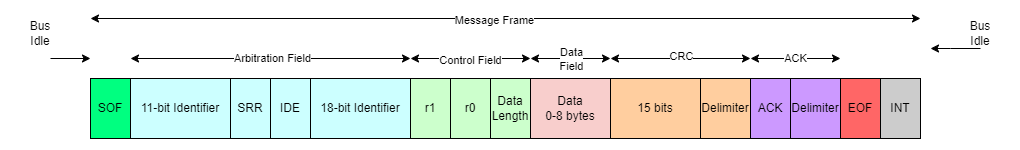}%
\label{fig_extended_can}}
\caption{Structure of CAN data frames. (a) Standard CAN 2.0A uses an 11-bit identifier. (b) Extended CAN 2.0B uses a 29-bit identifier, incorporating additional arbitration fields.}
\label{fig:fig_can_frames}
\end{figure*}

\subsection{Controller Area Network (CAN) }
Robert Bosch invented CAN, a serial communication protocol, in the early 1980s \cite{ref57}. Since it was invented, CAN has been used extensively to network ECUs in vehicles and in controlled systems, avoiding complicated wiring circuits. Figure~\ref{fig:fig_can_frames} shows two specifications of the CAN protocol: CAN 2.0A and CAN 2.0B.  In both cases, the CAN message starts with the Start of Frame (SOF) bit to indicate that the new frame has started. The arbitrator field differs between CAN 2.0A and CAN 2.0B, where CAN 2.0A accepts only 11-bit identifications and CAN 2.0B accepts both 11-bit and 29-bit identifications. On CAN 2.0A, the arbitration field contains an 11-bit identifier and then a Remote Transmission Request (RTR) bit to discriminate remote frames from data frames, while on CAN 2.0B there is an 11-bit identifier, Substitute Remote Request (SRR) bit, IDE bit followed by an 18-bit identifier which is used during 29-bit identifications. SRR behaves just like RTR for CAN 2.0A, while the IDE extension flag is specified if the code requires a 29-bit format. The control field contains two reserved bits (r1 and r0) to allow future expansion and a Data Length Code (DLC) to define the size of data in the data field. The data field contains the actual data being passed. Cyclic redundancy check (CRC) is a 15-bit code with a delimiter, which is present in both the specification and guarantees data integrity by finding faults in the transmitted data. After the CRC there is an acknowledgment (ACK) bit and then a delimiter, where the receivers confirm that the message has been received successfully. Finally, there is the End of the Field (EOF), which indicates the message ending, and the Intermission (INT) field for the bus to be inactive.

\subsection{Uncomplicated Application-level Vehicular Computing and Networking (UAVCAN) protocol}
UAVCAN is a lightweight protocol based on CAN 2.0B specification that offers secure communication for modern vehicles such as UAVs. It is a decentralized network, thereby avoiding a single point of failure. Nodes in the network can broadcast messages or invoke services to communicate with one another, which are uniquely identified by their numerical IDs. The UAVCAN protocol uses the principles of publish/subscribe for message broadcasting between multiple network nodes, permitting the transfer of actuator commands, status data, and sensor data. Moreover, for the service invocation, the protocol involves a two-step peer-to-peer process for node configuration and firmware updates. During the service invocation process, a server receives a request from a client node and processes it before responding to it. The data used in this study is defined based on UAVCAN protocol specifications.

\subsection{Graph-Based Models}
Graph-based models have been increasingly utilized to interpret data as graphs rather than in traditional tabular format, offering unique advantages in classification tasks. These models leverage the inherent relationships and structures within graph data, making them well-suited for various applications, including intrusion detection in UAVCAN networks. These models can be briefly categorized into transductive and inductive models.

\subsubsection{Transductive Models}
Transductive learning refers to when the model is trained and tested on the same graph, making predictions about nodes and edges that are present in the training graph. These models do not generalize to new nodes or graphs and are often confined to the graph used during training. The full graph structure is used during training, thus allowing the model to have better-learned embeddings (or representations) of the nodes within the same network. This often leads to better performance as the models have inherent knowledge about the entire graph and its nodes. However, transductive models cannot generalize to new nodes or graphs, thus making them less adaptable to dynamic graph scenarios, where new nodes or edges are frequently added or removed. One such example of a transductive model is GCNN.

The concept of Convolutional Neural Networks has been extended to graph-structured data by GCNNs. Convolution operations are carried out among the nodes of a graph, aggregating feature data from a node's neighbors to create new feature representations. This method works well when the underlying data distribution is greatly influenced by connection patterns or graph structure.

\subsubsection{Inductive Models}
Inductive learning refers to the scenario where models are trained on a portion of the graphs and then make predictions about nodes and edges previously unseen by the model. These models are beneficial when there is a sequence of dynamic graphs where nodes and edges are constantly added or removed. Because of their flexibility, inductive models can be evaluated on graphs different from the training graphs. GAT, GraphSAGE, and Graph-Based Transformer are inductive learning-based graph models that we have used.

GAT incorporates an attention mechanism into the graph convolution process to weigh the importance of neighboring nodes in the graph. This mechanism focuses on the most relevant information in a graph by giving attentional scores to nodes more relevant to the task performance, thus enhancing the ability to detect subtle anomalies.

GraphSAGE is an inductive learning framework that generates node embeddings by sampling and aggregating features from the node’s local neighborhood. This makes it suitable for dynamic and evolving network graphs like the ones generated in this study from CAN bus messages. By sampling a fixed number of neighbors compared to an entire neighborhood, GraphSAGE can be scaled to large graphs. Several different aggregation functions, such as mean, Long Short-Term Memory, or pooling, could be used to aggregate features of sampled neighborhoods, making it possible to capture different kinds of relationships in the graph.

Transformers were originally created for Natural Language Processing, which involves sequence-to-sequence tasks. Transformers have been modified for the use of graph-based data because of their capacity to detect long-range relationships via self-attention mechanisms. This enables the model to capture local and global structures by simultaneously considering the relationships between every pair of nodes in the graph. This capability makes the graph-based transformer powerful in detecting complex and long-range interactions that might indicate an intrusion.

\section{Dataset}
The Hacking and Countermeasure Research Lab (HCRL) provides the dataset used in this research \cite{ref27}. All data collection, testbed construction, and attack scenario designs were conducted by the HCRL team. A testbed was constructed using drones equipped with Pixhawk 4 (PX4) autopilot systems. The PX4 was connected to four Electronic Speed Controllers (ESCs) that control the motors via a serial CAN bus. The CAN bus was linked to Raspberry Pi 4, allowing for the reading, copying, and injecting of messages into the CAN bus. The decoded CAN bus messages—based on the UAVCAN protocol—were included as part of the released dataset, and no further protocol-level decoding was performed in this study. Various insertion attacks considered in this study are explained in the following subsections. As described in \cite{ref27}, a total of ten scenarios were defined based on different attack types and their corresponding settings, as presented in Table I.

\begin{table}[!t]
\caption{Summary of Attack Scenarios, Types, Number of Attacks, and Message Injection Interval\label{tab:attack_scenarios}}
\centering
\begin{tabular}{|P{1.5cm}|P{2.5cm}|P{1.5cm}|P{1.5cm}|}
\hline
\textbf{Scenario} & \textbf{Attack Type} & \textbf{Number of Attacks} & \textbf{Injection Interval (s)} \\
\hline
Scenario 1 & Flooding Attack & 3 & 0.0015 \\
\hline
Scenario 2 & Flooding Attack & 3 & 0.005 \\
\hline
Scenario 3 & Fuzzy Attack & 3 & 0.0015 \\
\hline
Scenario 4 & Fuzzy Attack & 3 & 0.005 \\
\hline
Scenario 5 & Replay Attack & 3 & 0.005 \\
\hline
Scenario 6 & Replay Attack & 4 & 0.005 \\
\hline
Scenario 7 & Flooding and Fuzzy Attack & 4 (2 from each) & 0.005 \\
\hline
Scenario 8 & Fuzzy and Replay Attack & 4 (2 from each) & 0.005 \\
\hline
Scenario 9 & Flooding and Replay Attack & 4 (2 from each) & 0.005 \\
\hline
Scenario 10 & Flooding, Fuzzy, and Replay Attack & 3 (1 from each) & 0.005 \\
\hline
\end{tabular}
\end{table}

\begin{figure*}[!t]
\centering
\includegraphics[width=0.8\textwidth]{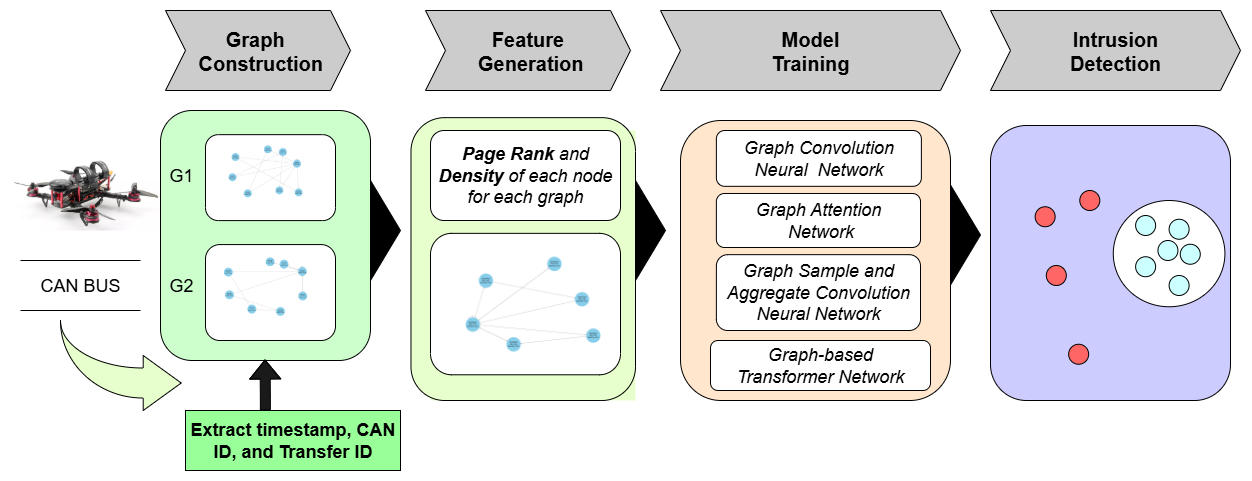}
\caption{Workflow of the proposed graph-based UAVCAN IDS}
\label{fig:gb_ids_workflow}
\end{figure*}

\subsection{Flooding attack}
A flooding attack is a Denial of Service (DoS) attack designed to consume resources on a server and prevent legitimate users and messages from using the resources. The dataset used in this study includes valid commands that halt the UAV's motors during the hovering phase. Scenarios 1 and 2 were created to collect data during flooding attacks. The only difference between these scenarios is the interval for injecting data, where scenario 1 uses an interval of 0.0015, and scenario 2 uses an interval of 0.005 seconds, as shown in Table I.

\subsection{Fuzzy attack}
A fuzzy attack injects random values into the message frame to induce abnormal behavior. The UAVCAN protocol uses a 29-bit identifier, which exempts the CAN ID field from being attacked directly with a random value generator. However, the shared data within the protocol are not protected by any security measures. Thus, random values can be inserted into the data of CAN messages, which may result in abnormal behavior of the UAVs. Table I shows Scenarios 3 and 4, designed for fuzzy attack, with injection intervals of 0.0015 and 0.005 seconds, respectively.

\subsection{Replay attack}
A replay attack, a prevalent and effective type of man-in-the-middle attack, involves copying and retransmitting communications while an attacker poses as a legitimate user. UAV directional control signals were pre-collected from the PX4 autopilot system. Table I shows Scenarios 5 and 6, specifically designed for replay attacks focusing on retransmitting left directional control messages at an injection interval of 0.005 seconds. 

\subsection{Mixed Attacks}
Pairwise combinations of flooding, fuzzy, and replay attacks were generated, as shown in Table I, with a data injection interval of 0.005 seconds. Scenario 10 included all three attacks, also with a data injection interval of 0.005 seconds.

\section{A Protocol-Independent Graph-Based Intrusion Detection Stratergy}
The graph-based IDS approach can be divided into four steps: (i) the graph construction module, (ii) the feature generation module, (iii) the model training module, and (iv) the intrusion detection module. In the graph construction and feature engineering module, we discuss the generation of graph streams from tabular time series data and calculating page rank features and density features. Our approach workflow is shown in Figure~\ref{fig:gb_ids_workflow}. We elaborate on data decoding, graph construction, feature engineering, model training, and intrusion detection in the following subsections.

% \begin{figure}[!t]
% \centering
% \includegraphics[width=\linewidth]{Workflow_GB_IDS.drawio.png}
% \caption{Workflow of the proposed graph-based UAVCAN IDS}
% \label{fig:gb_ids_workflow}
% \end{figure}

\subsection{Data Decoding}
\begin{figure*}[!t]
\centering
\includegraphics[width=0.9\textwidth]{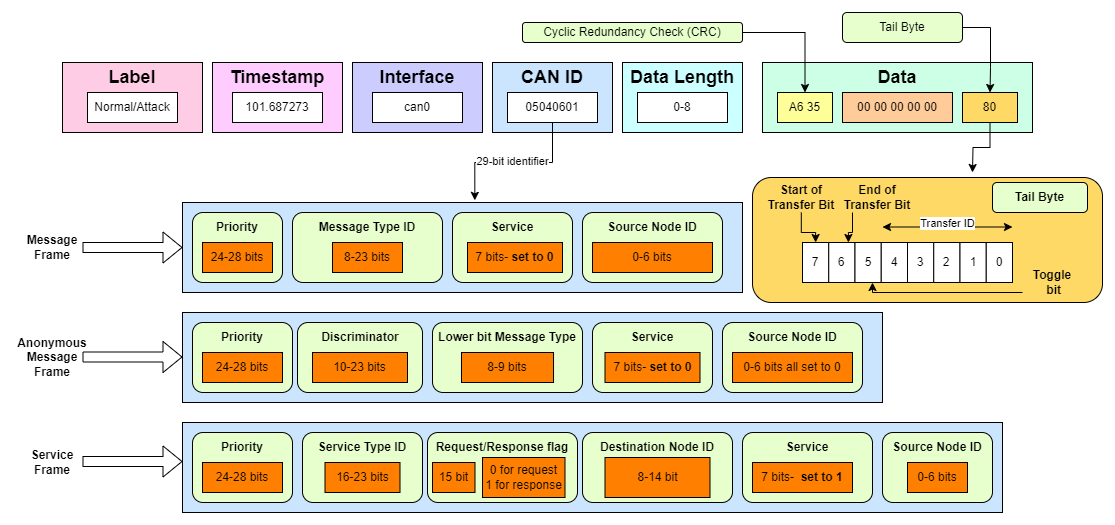}
\caption{Structured representation of UAVCAN messages derived from decoded CAN frames, showing bitwise composition across multiple frame types.}
\label{fig:uavcan_decoding}
\end{figure*}

The dataset consists of “.bin” files, from which columns like the label, timestamp, interface, CAN ID, data length, and data field were retrieved. As shown in Figure~\ref{fig:uavcan_decoding}, a hexadecimal tail byte of UAVCAN protocol messages includes encoded bits for the start and end of a transfer, a toggle bit, and a transfer ID \cite{ref27}. The remaining hexadecimal data from the data field was also decoded for single-frame and multi-frame messages. Based on the UAVCAN protocol, we ignored the first two hexadecimal bytes of the multi-frame message, which signify the Cyclic Redundancy Check (CRC) used for checking the data integrity of messages when targeted ECUs received all the frames of the multi-frame messages. The CAN ID from the UAVCAN protocol is a 29-bit identifier, which includes information like request or response message, service or non-service message, priority of the message, message type ID, and source node ID.

Following the decoding, the data was then cleaned, ensuring that the data type of each column was consistent with no null values. For our analysis during the graph construction phase, one feature, transfer ID, is used, which defines self-loops in graphs.

% \begin{figure}[!t]
% \centering
% \includegraphics[width=\linewidth]{decoding_pipeline.drawio.png}
% \caption{Data decoding based on UAVCAN protocol specification}
% \label{fig:uavcan_decoding}
% \end{figure}

\subsection{Graph Construction Module}
\begin{figure}[!t]
\centering
\includegraphics[width=\linewidth]{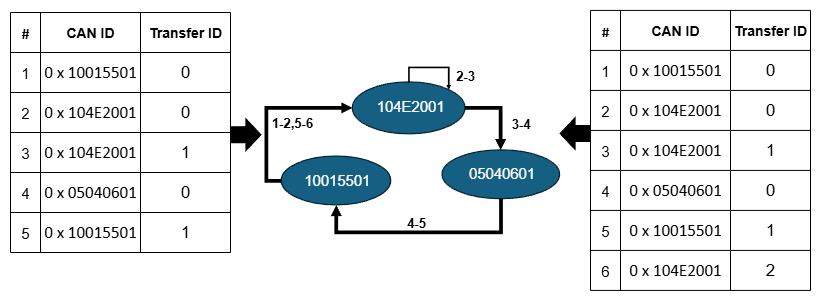}
\caption{Simple graph generation from a sequence of CAN IDs}
\label{fig:simple_graph_generation}
\end{figure}

This subsection discusses the process of converting decoded tabular CAN messages into graphs to identify intrusion behavior using minimal protocol specification. Only the \textit{Timestamp}, \textit{CAN ID}, and \textit{Transfer ID} were extracted for this task from the decoded dataset. CAN IDs were collected over a fixed message window $\Delta t$ and used to generate a stream of graphs $G_t$ at time $t_i$. To handle multi-frame messages, we incorporated a constraint where a self-loop is introduced between two consecutive identical CAN IDs only if there is a difference in the Transfer ID. This is because a difference in Transfer ID signifies a new message, as shown in the concept formulation in Equation~\ref{eq:1}. Figure~\ref{fig:simple_graph_generation} represents a simple graph depicting the concept formulation of Equation~\ref{eq:1} to ensure an accurate representation of message frames.

% \begin{equation}
% \label{eq:1}
% G_t = \bigcup_{t_i} \left( \text{CAN\_ID}_{t_i} \rightarrow \text{CAN\_ID}_{t_{i+1}} \right)
% \end{equation}

% \noindent\textbf{With constraint:}
% \[
% \text{if } \text{CAN\_ID}_{t_i} = \text{CAN\_ID}_{t_{i+1}} \text{ and }\text{Transfer\_ID}_{t_i} \ne \text{Transfer\_ID}_{t_{i+1}}
% \]
% \[
% t_i = \text{\textit{time definition window}} 
% \]
% \[
% t_i \in \left( t_0 + k \Delta t,\; t_0 + (k+1)\Delta t \right),\; k \in \mathbb{N}
% \]
\begin{equation}
\label{eq:1}
G_t = \bigcup_{t_i} \left( \text{CAN\_ID}_{t_i} \rightarrow \text{CAN\_ID}_{t_{i+1}} \right)
\end{equation}

% \vspace{-1em}
{\small
\begin{minipage}{\columnwidth}
\noindent With constraint: \\
\noindent
if $\text{CAN\_ID}_{t_i} = \text{CAN\_ID}_{t_{i+1}}$ and $\text{Transfer\_ID}_{t_i} \ne \text{Transfer\_ID}_{t_{i+1}}$ \\
$t_i$ = \textit{time definition window} \\
$t_i \in (t_0 + k\Delta t,\ t_0 + (k+1)\Delta t),\quad k \in \mathbb{N}$
\end{minipage}
}

\begin{figure}[!t]
\centering
\includegraphics[width=\linewidth]{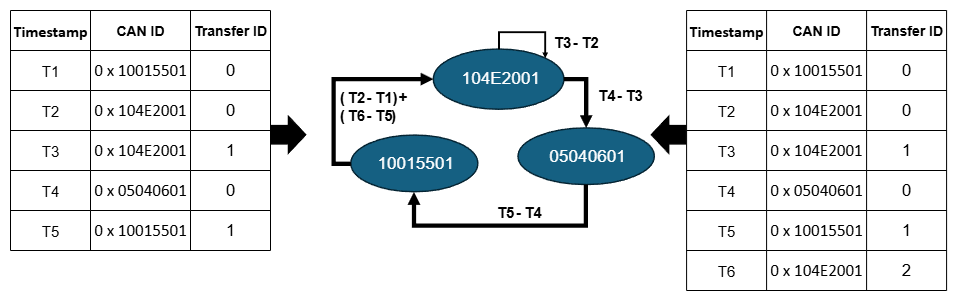}
\caption{An example of our graph construction approach.}
\label{fig:proposed_graph_design}
\end{figure}

Figure~\ref{fig:simple_graph_generation} illustrates a scenario where multiple transactions are represented as a single edge. To address this, we incorporate the sum of timestamp differences between two nodes as edge weights, as shown in Figure~\ref{fig:proposed_graph_design}. For this study, we used a fixed message window $\Delta t$ of 100 samples. Finally, we generated a stream of graphs from each $\Delta t$ window, denoted as $G_t = (V_t, E_t)$, where $V_t$ is the set of vertices and $E_t$ is the set of edges. Each graph contains $N_{V_t}$ vertices $\{V_1, V_2, \ldots, V_{N_{V_t}}\}$ and $N_{E_t}$ edges $\{E_1, E_2, \ldots, E_{N_{E_t}}\}$. Each vertex $V_i$ represents a unique CAN ID, such as $ID_i$ received at timestamp $TS_i$, and $ID_{i+1}$ at $TS_{i+1}$. The weight of each edge $E_i$ is defined as the sum of timestamp differences for all instances of the sequence within the $\Delta t$ window, as described in Equation~\ref{eq:2}. Figure~\ref{fig:proposed_graph_design} presents a modified version of the simple graph in Figure~\ref{fig:simple_graph_generation}, incorporating timestamp-based edge weights between two vertices.

\begin{equation}
\label{eq:2}
E_i = \left( V_{e(i)} \rightarrow V_{e(i+1)}, \sum_{i=1}^{100} (TS_{i+1} - TS_i) \right)
\end{equation}

\subsection{Feature Generation Module}
After the graph construction phase, a sequence of streams of graphs $G_t$ is obtained. The importance of each vertex in each graph is calculated using the PageRank algorithm and the density of occurrence in the past 150 samples is calculated along with the current window.
\subsubsection{PageRank Algorithm}
The Page Rank algorithm was developed by Google’s founders Larry Page and Sergey Brin and is used to rank websites for search engine optimization \cite{ref58}. Weights are assigned to each website, reflecting its importance through the outdegree, defined as the number of edges originating from the vertex. An iterative process is employed where rank values are distributed to linked pages based on current ranks until stabilization. 

Consider a graph with three nodes as shown in Figure~\ref{fig:simple_graph_generation}, $V_1 = 10E2001$, $V_2 = 10015501$, and $V_3 = 05040601$. Initially, an equal weight of $\frac{1}{N_{V_i}}$, where $N_{V_i}$ is the number of unique nodes in a time window (i.e., $\frac{1}{3}$ in this case), is assigned to ensure probabilities between 0 and 1. In each iteration, weights are recalculated based on the connected nodes as shown in Equation~\ref{eq:pagerank_basic}. Self-loops are avoided in PageRank calculation. A minimum value is set for isolated vertices to ensure each node can be visited with minimal probability. We used $d$ as a damping factor and modified the PageRank equation as shown in Equation~\ref{eq:pagerank_damping}. To include the edge weights in our calculation, we have optimized the PageRank formula in Equation~\ref{eq:pagerank_basic} by dividing it with a summation of the edge weights between $V_1$ and the respective nodes, as shown in Equation~\ref{eq:pagerank_weighted}.

\begin{equation}
\label{eq:pagerank_basic}
\begin{split}
\text{PageRank}(V_1) = 
\frac{\text{PageRank}(V_2)}{\text{OutDegree}(V_2)} + 
\frac{\text{PageRank}(V_3)}{\text{OutDegree}(V_3)}
\end{split}
\end{equation}

\begin{equation}
\label{eq:pagerank_damping}
\begin{split}
\text{PageRank}(V_1) = \left( \frac{1 - d}{N_{V_i}} \right) 
+ d \Bigg( 
\frac{\text{PageRank}(V_2)}{\text{OutDegree}(V_2)} \\
+ \frac{\text{PageRank}(V_3)}{\text{OutDegree}(V_3)} 
\Bigg)
\end{split}
\end{equation}

% \begin{align}
% \label{eq:pagerank_weighted}
% \text{PageRank}(V_1) &= \left( \frac{1 - d}{N_{V_i}} \right) \nonumber \\
%  + d \left( 
% \sum_{p_i \in \{V_2, V_3\}} 
% \frac{
% \text{PageRank}(V_2) / \text{OutDegree}(V_2)
% }{
% \text{EdgeWeight}(V_1, p_i)
% }
% \right)
% \end{align}

\begin{align}
\label{eq:pagerank_weighted}
\text{PageRank}(V_1) &= \left( \frac{1 - d}{N_{V_i}} \right) \nonumber \\
&\quad + d \left( 
\sum_{p_i \in \{V_2, V_3\}} 
\frac{
\frac{\text{PageRank}(p_i)}{\text{OutDegree}(p_i)}
}{
\text{EdgeWeight}(V_1, p_i)
}
\right)
\end{align}

\subsubsection{Density Feature}
\begin{figure}[!t]
    \centering
    \includegraphics[width=0.45\textwidth]{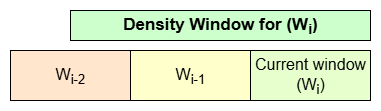}
    \caption{Illustration of the density window used for calculating the local density for current window $W_i$}
    \label{fig:density_window}
\end{figure}

The density of each node, i.e., CAN ID, is calculated with the current window and 150 samples that preceded the current window, as shown in Figure~\ref{fig:density_window}. Density is one of the protocol-independent parameters we choose as a feature for our GB-ML models. This enables us to extend these models to other CAN-based protocols by just considering the CAN ID and its density.

\subsection{Model Training and Intrusion Detection Module}
Model training involves incorporating class weights for each class to address data imbalance during the training phase. Class weights are added to avoid overfitting the model to a larger number of non-attack instances, as the non-attack instances in our dataset outnumber the attack instances. Thus making sure that the model learns to identify both attack and non-attack data. Once trained, the models were evaluated to assess the IDS’s effectiveness across a variety of attack scenarios.

% Model training involves two key steps: incorporating class weights for each class in the training phase and using different graph-based models to test the detection capabilities of the IDS for UAVCAN networks. Class weights were considered during the training phase to address the data imbalance issue, where non-attack instances outnumber attack instances. This is crucial as it prevents the model from becoming biased towards non-attack data instances, which can lead to poor performance during attack instances. Therefore, more priority was assigned to attack instances during the training phase, assuring the model learns to recognize both attack and non-attack data appropriately.

This study evaluated four different graph-based models to leverage the spatial and temporal information in UAVCAN network data, such as GCNN, GAT, GraphSAGE, and Graph-Based Transformer. Metrics like accuracy, precision, recall, and F1 scores were used to evaluate the models during each type of attack. The goal was to achieve higher performance accuracy during individual and mixed attack scenarios.

\section{RESULTS AND DISCUSSION}
The IDS performance was evaluated using inductive and transductive learning models to leverage the spatial and temporal information in CAN bus network data. The study evaluated four graph-based machine-learning models, namely GCNN, GAT, GraphSAGE, and Graph-based Transformer, to assess their detection performance when the CAN bus network of UAVs is attacked. Tables~\ref{tab:graphsage_accuracy} and ~\ref{tab:transformer_accuracy} represent the evaluation metrics for GraphSAGE and Graph-Based transformer models generated across ten attack scenarios, respectively.

\begin{table}[!t]
\caption{Detection Performance for GraphSAGE Model}
\label{tab:graphsage_accuracy}
\centering
\renewcommand{\arraystretch}{1.2}
\begin{tabular}{|c|c|c|c|c|}
\hline
\textbf{Scenario} & \textbf{Accuracy (\%)} & \textbf{Precision} & \textbf{Recall} & \textbf{F1-Score} \\
\hline
Scenario 1 & 0.994 & 0.990 & 0.994 & 0.992 \\
Scenario 2 & 0.979 & 0.511 & 0.740 & 0.516 \\
Scenario 3 & 0.988 & 0.976 & 0.986 & 0.981 \\
Scenario 4 & 0.997 & 0.500 & 0.499 & 0.499 \\
Scenario 5 & 0.836 & 0.556 & 0.887 & 0.556 \\
Scenario 6 & 0.929 & 0.788 & 0.952 & 0.843 \\
Scenario 7 & 0.910 & 0.764 & 0.941 & 0.818 \\
Scenario 8 & 0.935 & 0.832 & 0.946 & 0.876 \\
Scenario 9 & 0.810 & 0.532 & 0.855 & 0.507 \\
Scenario 10 & 0.909 & 0.711 & 0.919 & 0.768 \\
\hline
\end{tabular}
\end{table}

The performance of GraphSAGE and Graph-Based transformers was highly effective in detecting various attack scenarios for CAN bus networks in UAVs. As an inductive graph model, these models utilized mean aggregation to achieve results with high precision, recall, and F1-score as shown in Tables~\ref{tab:graphsage_accuracy} and ~\ref{tab:transformer_accuracy}. 

\begin{table}[!t]
\caption{Detection Performance for Graph-based Transformer Model}
\label{tab:transformer_accuracy}
\centering
\renewcommand{\arraystretch}{1.2}
\begin{tabular}{|c|c|c|c|c|}
\hline
\textbf{Scenario} & \textbf{Accuracy (\%)} & \textbf{Precision} & \textbf{Recall} & \textbf{F1-Score} \\
\hline
Scenario 1 & 0.994 & 0.990 & 0.994 & 0.992 \\
Scenario 2 & 0.967 & 0.507 & 0.734 & 0.505 \\
Scenario 3 & 0.985 & 0.971 & 0.982 & 0.976 \\
Scenario 4 & 0.999 & 0.500 & 0.500 & 0.500 \\
Scenario 5 & 0.843 & 0.559 & 0.898 & 0.562 \\
Scenario 6 & 0.945 & 0.818 & 0.960 & 0.871 \\
Scenario 7 & 0.917 & 0.773 & 0.945 & 0.828 \\
Scenario 8 & 0.934 & 0.830 & 0.944 & 0.873 \\
Scenario 9 & 0.816 & 0.533 & 0.858 & 0.511 \\
Scenario 10 & 0.899 & 0.699 & 0.922 & 0.753 \\
\hline
\end{tabular}
\end{table}
Moreover, Table~\ref{tab:gcnn_accuracy} and ~\ref{tab:gat_accuracy} present the evaluation metrics for the GCNN and GAT models. However, it performed moderately compared to GraphSAGE and Graph-based transformers. As discussed earlier, the GCNN model, being a transductive model, failed to scale out to samples beyond the training graph, which resulted in lower performance in a few attack scenarios. The GAT model, although an inductive model, did not achieve the same level of performance as GraphSAGE and Graph-based transformers since it leverages local neighborhood information without being able to effectively capture the global graph structure, which is essential for identifying complex attacks.

\begin{table}[!t]
\caption{Detection Performance for GCNN Model}
\label{tab:gcnn_accuracy}
\centering
\renewcommand{\arraystretch}{1.2}
\begin{tabular}{|c|c|c|c|c|}
\hline
\textbf{Scenario} & \textbf{Accuracy (\%)} & \textbf{Precision} & \textbf{Recall} & \textbf{F1-Score} \\
\hline
Scenario 1 & 0.910 & 0.879 & 0.907 & 0.891 \\
Scenario 2 & 0.971 & 0.500 & 0.486 & 0.493 \\
Scenario 3 & 0.861 & 0.780 & 0.864 & 0.808 \\
Scenario 4 & 1.000 & 0.500 & 0.500 & 0.500 \\
Scenario 5 & 0.799 & 0.541 & 0.817 & 0.521 \\
Scenario 6 & 0.852 & 0.690 & 0.889 & 0.730 \\
Scenario 7 & 0.863 & 0.695 & 0.846 & 0.734 \\
Scenario 8 & 0.901 & 0.776 & 0.883 & 0.815 \\
Scenario 9 & 0.783 & 0.525 & 0.803 & 0.488 \\
Scenario 10 & 0.860 & 0.659 & 0.899 & 0.701 \\
\hline
\end{tabular}
\end{table}

To evaluate the performance of the GB-ML models further, we compare them with a baseline single-layer LSTM model, as shown in Table~\ref{tab:comparison_accuracy}. Unlike the graph-based models that rely on graph-structured features such as PageRank and density, the LSTM model was trained on decoded CAN data derived directly from the UAVCAN protocol, as shown in \cite{ref59}. This includes features extracted after parsing the raw binary logs, such as timestamp, data length, and data fields, but without any graph-based transformation or connectivity representation. In most cases, the LSTM model consistently performed worse than the graph-based models.  For example, in Scenario 1, the LSTM obtained an accuracy of 69.81\%, which was poor compared with either GraphSAGE’s or the Graph-Based Transformer’s accuracy of 99.45\%. The LSTM could not match the performance of the graph-based models, even in Scenario 4, when it performed at its peak accuracy of 93.2\%. GB-ML models consistently outperformed the LSTM in more complex attack scenarios, such as scenarios 6 to 10. These results show that while LSTMs are able to capture temporal sequences, graph-based models better capture the complexity of the communication structures within CAN scenarios, leading to superior performance in the intrusion detection task.

\begin{table}[!t]
\caption{Detection Performance for GAT Model}
\label{tab:gat_accuracy}
\centering
\renewcommand{\arraystretch}{1.2}
\begin{tabular}{|c|c|c|c|c|}
\hline
\textbf{Scenario} & \textbf{Accuracy (\%)} & \textbf{Precision} & \textbf{Recall} & \textbf{F1-Score} \\
\hline
Scenario 1 & 0.881 & 0.845 & 0.896 & 0.862 \\
Scenario 2 & 0.976 & 0.500 & 0.488 & 0.494 \\
Scenario 3 & 0.790 & 0.728 & 0.846 & 0.742 \\
Scenario 4 & 0.998 & 0.500 & 0.499 & 0.500 \\
Scenario 5 & 0.844 & 0.537 & 0.730 & 0.531 \\
Scenario 6 & 0.697 & 0.609 & 0.796 & 0.585 \\
Scenario 7 & 0.800 & 0.648 & 0.821 & 0.670 \\
Scenario 8 & 0.942 & 0.854 & 0.911 & 0.879 \\
Scenario 9 & 0.794 & 0.521 & 0.741 & 0.486 \\
Scenario 10 & 0.892 & 0.681 & 0.870 & 0.729 \\
\hline
\end{tabular}
\end{table}
Table~\ref{tab:comparison_accuracy} presents a comprehensive comparison of detection accuracy across all ten attack scenarios. It is evident that both GraphSAGE and Graph-Based Transformer models maintain high detection accuracy throughout, with values consistently above 90\% in most scenarios. In contrast, the LSTM model shows notable drops in performance, especially in mixed and complex attack scenarios such as Scenario 8 (62.6\%) and Scenario 10 (68.6\%). GCNN and GAT models perform moderately well but still fall short of the accuracy achieved by GraphSAGE and Graph-Based Transformer.

Overall, the GB-ML models demonstrated improvements ranging from approximately 1\% to 29\% over the LSTM model across all attack scenarios, highlighting their robustness and adaptability in varied intrusion contexts. These observations indicated that graph-based methods could potentially have a promising role in enhancing the IDS's performance in CAN bus networks.

\begin{table}[!t]
\caption{Comparison of Detection Accuracy of GB-ML models with LSTM model}
\label{tab:comparison_accuracy}
\centering
\renewcommand{\arraystretch}{1.2}
\begin{tabular}{|P{1.5cm}|P{1.5cm}|P{0.8cm}|P{0.8cm}|P{0.8cm}|P{0.8cm}|}
\hline
\textbf{Scenario} & \textbf{GraphSAGE (\%)} & \textbf{Graph-based Transformer (\%)} & \textbf{GCNN (\%)} & \textbf{GAT (\%)} & \textbf{LSTM (\%)} \\
\hline
Scenario 1 & 0.994 & 0.994 & 0.910 & 0.881 & 0.698 \\
Scenario 2 & 0.979 & 0.967 & 0.971 & 0.976 & 0.820 \\
Scenario 3 & 0.988 & 0.985 & 0.861 & 0.790 & 0.820 \\
Scenario 4 & 0.997 & 0.999 & 1.000 & 0.998 & 0.932 \\
Scenario 5 & 0.836 & 0.843 & 0.799 & 0.844 & 0.812 \\
Scenario 6 & 0.929 & 0.945 & 0.852 & 0.697 & 0.706 \\
Scenario 7 & 0.910 & 0.917 & 0.863 & 0.800 & 0.779 \\
Scenario 8 & 0.935 & 0.934 & 0.901 & 0.942 & 0.626 \\
Scenario 9 & 0.810 & 0.816 & 0.783 & 0.794 & 0.612 \\
Scenario 10 & 0.909 & 0.899 & 0.860 & 0.892 & 0.686 \\
\hline
\end{tabular}
\end{table}

\section{CONCLUSIONS}
In this study, we developed and evaluated a graph machine learning-based IDS for UAVs communicating via the UAVCAN protocol. We encoded UAVCAN protocol messages into graphs and used them as input for various GB-ML models. These models demonstrate higher detection performance from 1\% to 29\% compared to the baseline LSTM-based IDS, in terms of precision, recall, and F1 scores in detecting various injection attacks on the CAN bus network.

The results show that graph-based methods outperform conventional sequence-based models like LSTMs in detecting intrusions because they are highly effective at capturing intricate communication structures of CAN bus data. Therefore, graph-based approaches are likely better at identifying intrusions, particularly in mixed or more sophisticated attacks. Moreover, the IDS models described in this study offer a flexible solution that can be applied to any UAV platform developed by different UAV manufacturers since these models depend on the density and sequence of data transferred in the communication channel rather than the protocol specification. Even though the models produce encouraging results, further work is required to optimize these systems for real-time deployment, particularly for UAVs with limited resources.  To obtain more robust results, we need to test these systems in a real-time operational condition and study the impact on a UAV’s processor capacity and latency.

In addition, our analysis demonstrates that the graph-based IDS has significant potential in the market for consumer, commercial and military-grade UAVs. For consumer-grade UAVs, lightweight variants of graph-based models such as GraphSAGE can provide security by leveraging their capability to capture communication patterns with relatively low computational overhead. On the other hand, commercial and military-grade UAVs, where security requirements are more stringent and attacks are more complex, could benefit from the full capabilities of these graph-based models, like graph-based transformers. The model's performance in detecting complex mixed attack scenarios indicates its ability to detect sophisticated attack events in higher-risk contexts.

\section{FUTURE SCOPE}
In our future work, we will optimize the graph-based IDS for deployment in UAVs to detect intrusions in real-time while evaluating its performance in various operational environments with different computational capacities. Deploying these models in different UAV platforms, from reasonably priced consumer-grade UAVs to more expensive military-grade UAVs, will allow us to evaluate the scalability and adaptability of this approach for real-world usage. Such real-world tests will also allow us to fine-tune the models to ensure acceptable low-latency performance with minimal computational overhead that can be accommodated by various UAVs.

Moreover, a UAV trajectory depends on the mission's goal, which determines the direction, speed, and altitude needed. For example, search and rescue operations require maximum coverage, while surveillance operations require stealth to avoid detection \cite{ref60}. Mission requirements like engaging with a target or avoiding obstacles during a flight can define intricate courses of action executed by exchanging information via a UAV’s CAN bus. Deviation and disturbances during these flight operations can compromise flight objectives, thus highlighting the necessity for trajectory features for mission effectiveness and safety. Based on these aspects, in future research, we will consider using trajectory-based features in our IDS models to support intent-based anomaly detection. Trajectory-aware IDS can detect deviations from the mission path using intention-based trajectory patterns of the UAV to identify potential attacks and system faults without compromising consistency and mission completion, security and trajectory limitations.

 Additionally, we would like to include explicit timing-based anomaly detection mechanisms that could be incorporated with graph features to detect timing abnormalities in CAN bus networks. This would enable us to test the performance-related benefits of adding explicit timing thresholds to our graph-based features to better detect attacks, unexpected input patterns, and latency-related problems.

 Moreover, using a graph-based framework for other CAN-based protocols could help us improve the applicability of this approach beyond UAVs to a wide variety of vehicular communication systems. Furthermore, we would also investigate if graph-based models can be used with LSTM aggregation to enhance the temporal learning capability and the overall accuracy of our IDS.

\section*{Acknowledgments}
This work is based upon the work partially supported by the National Center for Transportation Cybersecurity and Resiliency (TraCR) (a U.S. Department of Transportation National University Transportation Center) and by the Department of Energy Minority Serving Institutions Partnership Program (MSIPP) managed by the Savannah River National Laboratory under BSRA contract TOA0000525174 CN1. Any opinions, findings, conclusions, and recommendations expressed in this material are those of the author(s). They do not necessarily reflect the views of TraCR or the Savannah River National Laboratory, and the U.S. Government assumes no liability for the contents or use thereof.

% \newpage
% \vspace*{-5\baselineskip} 
\section*{Biography Section}
\vspace*{-3.5\baselineskip} % Pull the section closer

\begin{IEEEbiography}[{\includegraphics[width=1in,height=1.25in,clip,keepaspectratio]{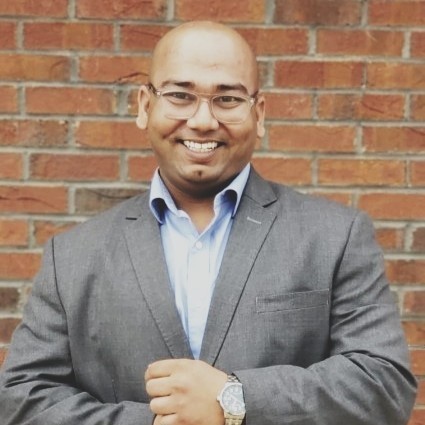}}]{Reek Majumder}
received his B.Tech. in Computer Science from KIIT, Bhubaneswar in 2015, followed by industry roles at Cognizant, where he worked as a software developer on automated testing and NLP-based Quality tools. He earned his M.Sc. in Computer Science (Data Science and Cybersecurity) and a Ph.D. in Civil Engineering (Transportation Systems) under the supervision of Dr. Mashrur Chowdhury from Clemson University. His research focuses on Connected and Autonomous Vehicles (CAVs) and Unmanned Aerial Vehicles (UAVs), with core interests in machine learning, quantum machine learning, graph-based learning, Agentic AI, and cybersecurity.
\end{IEEEbiography}
\vspace*{-3\baselineskip}

\begin{IEEEbiography}[{\includegraphics[width=1in,height=1.25in,clip,keepaspectratio]{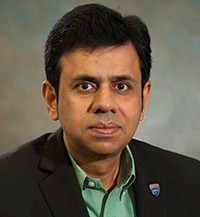}}]{Mashrur “Ronnie” Chowdhury}
is the Eugene Douglas Mays Professor and Chair of Transportation at Clemson University. He directs the USDOT Center for Connected Multimodal Mobility (C2M2), National Center for Transportation CyberSecurity and Resiliency (TraCR) and co-directs the Complex Systems, Analytics, and Visualization Institute (CSAVI). He also leads the Transportation Cyber-Physical Systems Laboratory at Clemson. A Fellow of ASCE and Senior Member of IEEE, Dr. Chowdhury has served on the IEEE ITS Society Board of Governors and is an alumnus of the NAE Frontiers of Engineering program. He is a member of the TRB committees on Artificial Intelligence and ITS and the founding advisor of the IEEE ITSS Student Chapter at Clemson University. He is a registered professional engineer in Ohio.
\end{IEEEbiography}
\vspace*{-3\baselineskip}

\begin{IEEEbiography}[{\includegraphics[width=1in,height=1.25in,clip,keepaspectratio]{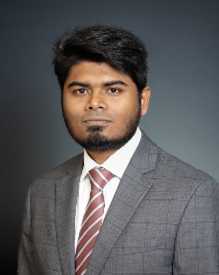}}]{M. Sabbir Salek}
(Member, IEEE) received his Ph.D. and M.S. in Civil Engineering from Clemson University in 2023 and 2021, respectively, and his B.S. in Mechanical Engineering from BUET, Dhaka, in 2016. He is a Senior Engineer for the USDOT-supported National Center for Transportation Cybersecurity and Resiliency (TraCR) and a Senior Research Engineer for the Center for Regional and Rural Connected Communities (CR2C2). He is also Adjunct Faculty at Clemson University. His research interests include cybersecurity and resiliency in intelligent transportation systems (ITS), connected and autonomous vehicles, and high-performance computing for ITS applications.
\end{IEEEbiography}

\vspace*{-3\baselineskip}
\begin{IEEEbiography}[{\includegraphics[width=1in,height=1.25in,clip,keepaspectratio]{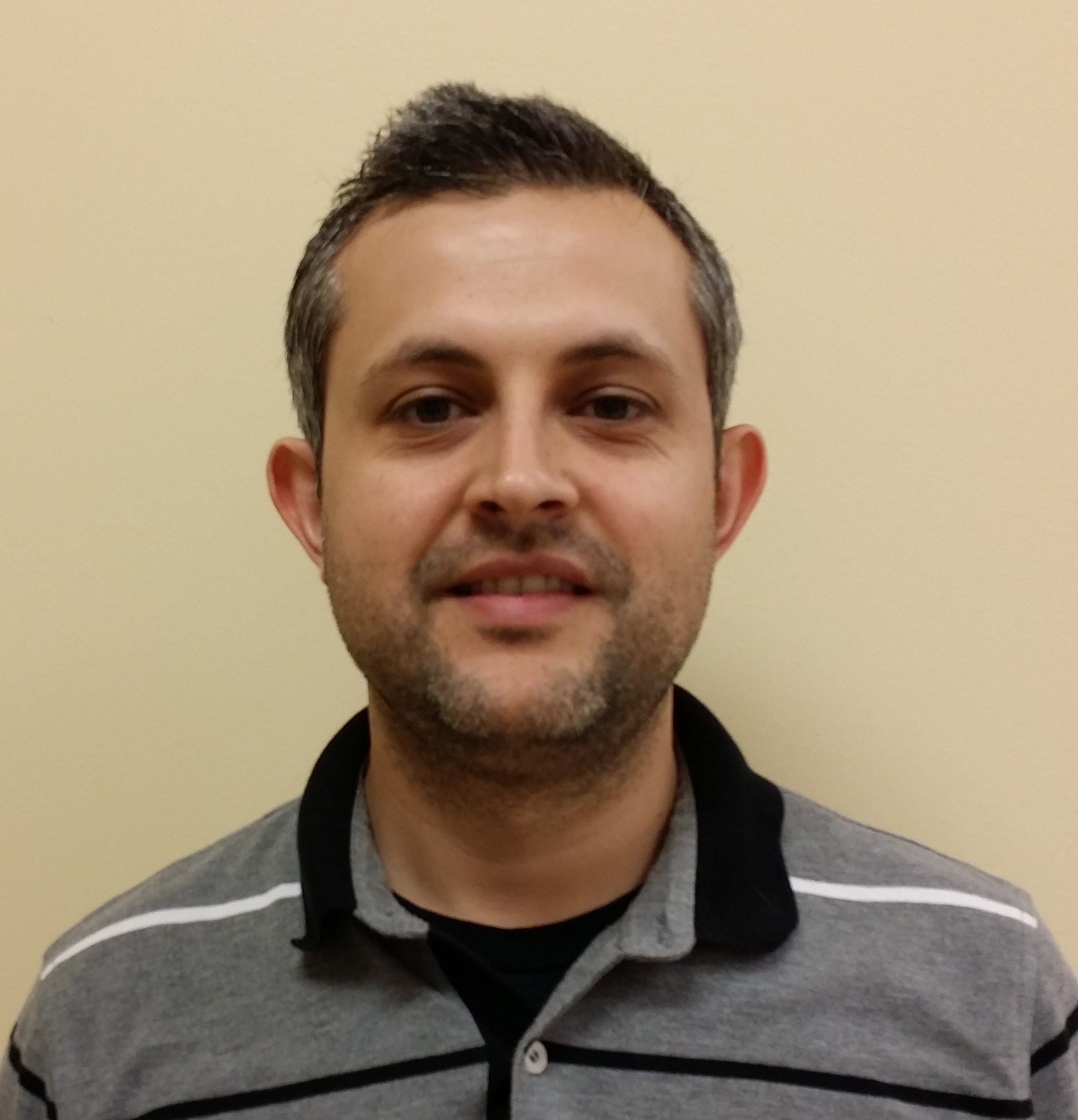}}]{Gurcan Comert}
received the B.Sc. and M.Sc. degrees in Industrial Engineering from Fatih (Istanbul) University, Istanbul, Turkey, and the Ph.D. degree in Civil Engineering from the University of South Carolina, Columbia, SC, in 2003, 2005, and 2008, respectively. He is currently with the Computational Data Science and Engineering Department at North Carolina A\&T State University, Greensboro, NC. His research interests include the modeling of various transportation problems.
\end{IEEEbiography}

\vspace*{-3\baselineskip}
\begin{IEEEbiography}[{\includegraphics[width=1in,height=1.25in,clip,keepaspectratio]{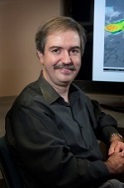}}]{David Werth}
joined the Atmospheric Technologies Group (ATG) of the Savannah River National Laboratory (SRNL) in 2006.  His research focuses on the mesoscale climate and airborne dispersion modeling.  He has participated in the study of evolutionary programming methods to improve mesoscale models and a study of the nocturnal boundary layer and low-level jet over Oklahoma.  He has also been involved in carbon monitoring at the Savannah River Site, the statistical downscaling of global climate model data as applied to maintain site operations, and become involved in the use of analog forecasting to predict available solar energy.
\end{IEEEbiography}

\vspace*{-2.5\baselineskip}
\begin{IEEEbiography}[{\includegraphics[width=1in,height=1.25in,clip,keepaspectratio]{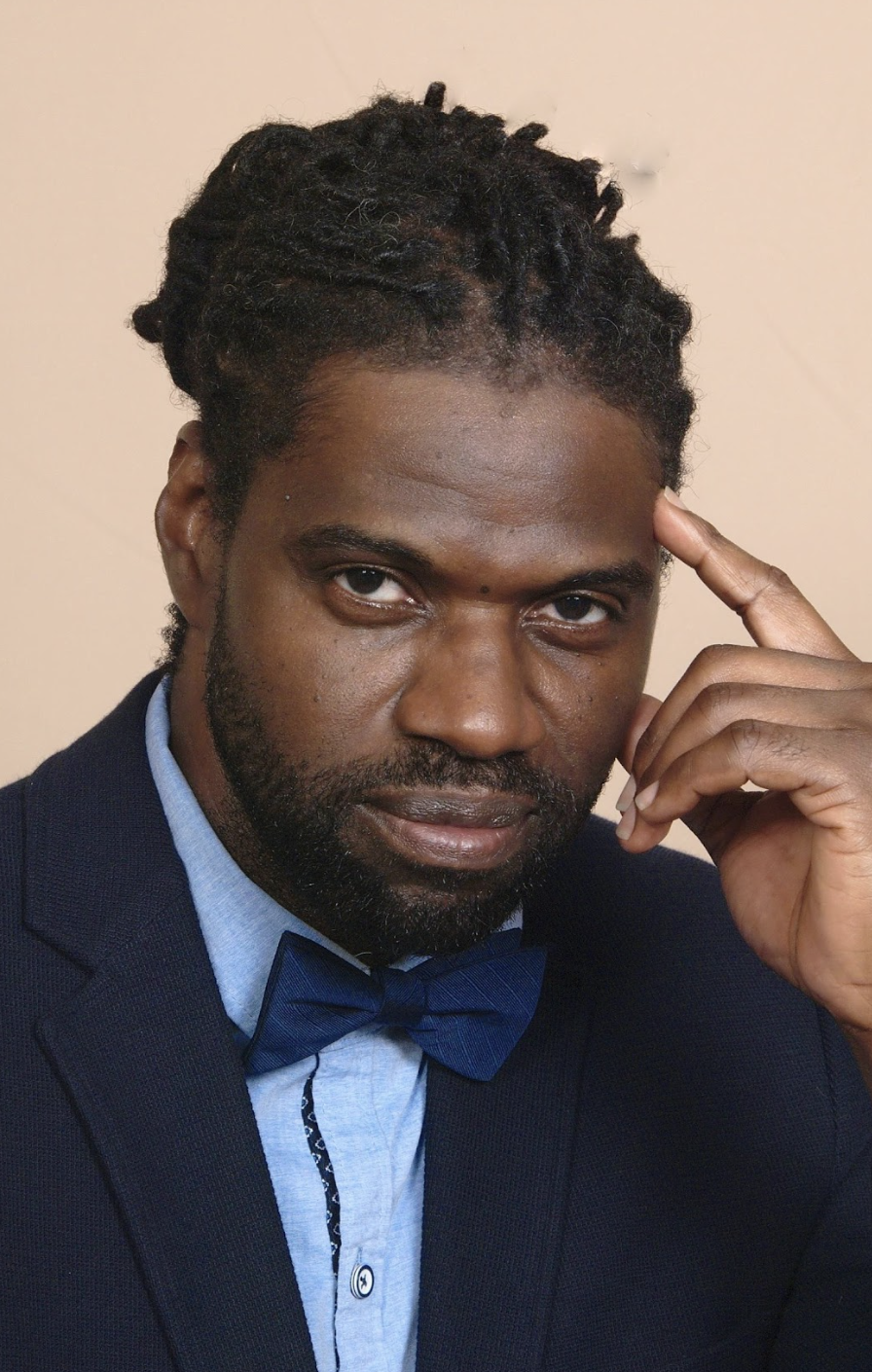}}]{Adrian Gale}
received his B.S.  degree from South Carolina State University, Orangeburg, SC, in Civil Engineering, M.S. Degree from the University of Illinois at Urbana-Champaign, Urbana, IL, in Environmental Engineering, and PhD from the University of Florida, Gainesville, FL, in Environmental Engineering in 2003, 2005, and 2013, respectively. He is currently an Assistant Professor of Environmental Engineering at Benedict College. Adrian’s research focuses on vulnerabilities in earth and environmental systems connected to legacy and emerging impacts from natural and anthropogenic activities (contamination, climate change, and post-disaster waste management). His work also links sustainability to urban ecosystem restoration and environmental justice.
\end{IEEEbiography}



\vfill

\end{document}